\documentclass [sigconf]{acmart}
\AtBeginDocument{%
  }

\setcopyright{acmlicensed}
\copyrightyear{2018}
\acmYear{2018}
\acmDOI{XXXXXXX.XXXXXXX}
\acmConference[Conference acronym 'XX]{Make sure to enter the correct
  conference title from your rights confirmation email}{June 03--05,
  2018}{Woodstock, NY}
\acmISBN{978-1-4503-XXXX-X/2018/06}
\usepackage[table]{xcolor}
\usepackage{booktabs}
\usepackage{multirow}
\usepackage{enumitem}
\usepackage{hyperref}
\usepackage{pifont} 
\setcopyright{none}

\renewcommand{\footnotetextcopyrightpermission}[1]{}

\begin{document}

\title{Model the Annotation, Not the Noise: Semantic Annotation-Channel Prompt Learning against Noisy Labels}

\settopmatter{authorsperrow=3}
\author{Zibin Geng}
\authornote{Both authors contributed equally to this research.}
\affiliation{%
  \institution{Institute of Computing Technology, Chinese Academy of Sciences}
  \institution{University of Chinese Academy of Sciences}
  \city{Beijing}
  \country{China}}
\email{gengzibin25z@ict.ac.cn}

\author{Yitong Bao}
\authornotemark[1]
\affiliation{%
  \institution{University of International Business and Economics}
  \city{Beijing}
  \country{China}}

\author{Jia Li}
\affiliation{%
  \institution{University of Chinese Academy of Sciences}
  \city{Beijing}
  \country{China}}

\author{Xuefeng Jiang}
\affiliation{%
  \institution{University of Chinese Academy of Sciences}
  \city{Beijing}
  \country{China}}

\author{Sheng Sun}
\affiliation{%
  \institution{Institute of Computing Technology, Chinese Academy of Sciences}
  \city{Beijing}
  \country{China}}

\author{Yuwei Wang}
\affiliation{%
  \institution{Institute of Computing Technology, Chinese Academy of Sciences}
  \city{Beijing}
  \country{China}}

\author{Min Liu}
\authornote{Corresponding author.}
\affiliation{%
  \institution{Institute of Computing Technology, Chinese Academy of Sciences}
  \city{Beijing}
  \country{China}}
\email{liumin@ict.ac.cn}

\renewcommand{\shortauthors}{Trovato et al.}

\begin{abstract}
Parameter-efficient prompt learning has become a cornerstone for adapting pre-trained vision-language models (VLMs), yet it remains highly vulnerable to label noise in few-shot Web and retrieval scenarios. Existing noise robust methods typically rely on hard sample selection or deterministic label correction, which discard valuable semantic ambiguity and overlook how observed annotations are actually generated. To bridge this gap, we reformulate noisy prompt learning through the perspective of data generation, proposing \texttt{SACP}, a \underline{S}emantic-aware \underline{A}nnotation \underline{C}hannel-based \underline{P}rompt learning framework. Instead of suppressing noise heuristically, \texttt{SACP} decouples latent clean-label prediction from instance-dependent annotation generation. By integrating frozen VLM semantic priors with lightweight visual residuals, \texttt{SACP} constructs a structured annotation channel optimized via generalized expectation-maximization with soft posteriors. Furthermore, a posterior-aware memory is introduced to preserve label uncertainty during test-time inference. 
Extensive experiments across different synthetic benchmarks and the real-world noisy Food101N dataset demonstrate that \texttt{SACP} achieves state-of-the-art performance against its counterparts in different label noise patterns, especially under severe asymmetric noise.
\end{abstract}

\ccsdesc[500]{Computing methodologies~Machine learning}
\ccsdesc[500]{Computing methodologies~Supervised learning by classification}
\ccsdesc[500]{Computing methodologies~Machine learning approaches}
\keywords{
Vision-language models,
prompt learning,
noisy-label learning,
few-shot learning,
annotation noise modeling
}


\maketitle

\section{Introduction}
\begin{figure}[htbp]
    \centering
    \includegraphics[width=\linewidth]{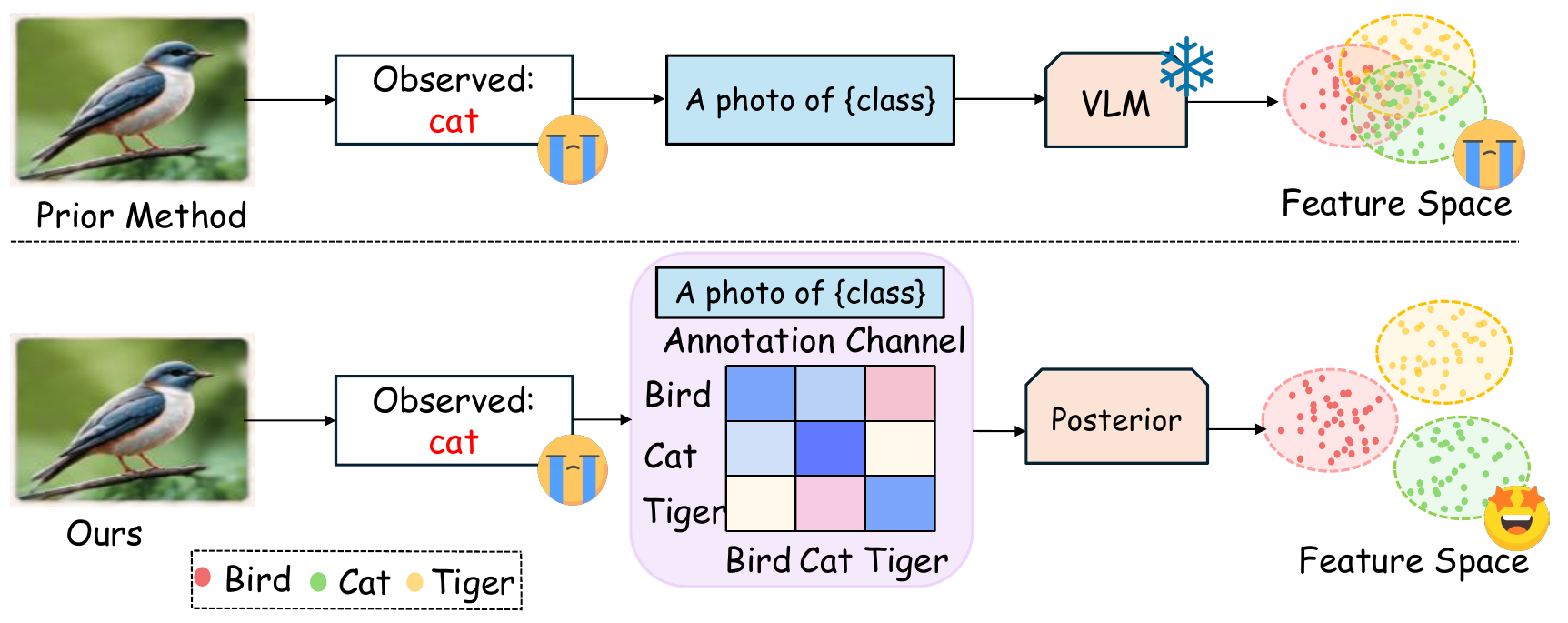}
    \caption{
Illustration of \texttt{SACP}.
Directly learning from noisy annotations may cause
semantic misalignment.
In contrast, \texttt{SACP} models how observed annotations
arise from latent clean classes and infers clean-label
posteriors to improve semantic alignment.
}
    \label{fig:teaser}
\end{figure}

Vision-language models (VLMs)
\cite{CLIP,jia2021scaling,li2025surveystateartlarge}
learn transferable representations from large-scale image-text
pairs, enabling strong zero-shot recognition and retrieval \cite{blip,force,tbd}. For zero-shot classification, these models match image features
with handcrafted text prompts, such as
``a photo of a [CLASS]'', by embedding images
and text into a shared semantic space \cite{CLIP,zero}.
However, manually designed prompts are sensitive to wording
and require substantial effort \cite{manu,manu2}.
Therefore, prompt learning has emerged as an efficient
and robust approach to adapting VLMs \cite{CoOp,cocoop,nlprompt,geng2026seeing}. Benefiting from a small number of trainable parameters
(typically only a few thousand), prompt learning enables
rapid optimization of prompt parameters through
backpropagation \cite{plot,fpl} for efficient downstream adaptation.
\begin{figure*}[htbp]
    \centering
    \includegraphics[width=\linewidth]{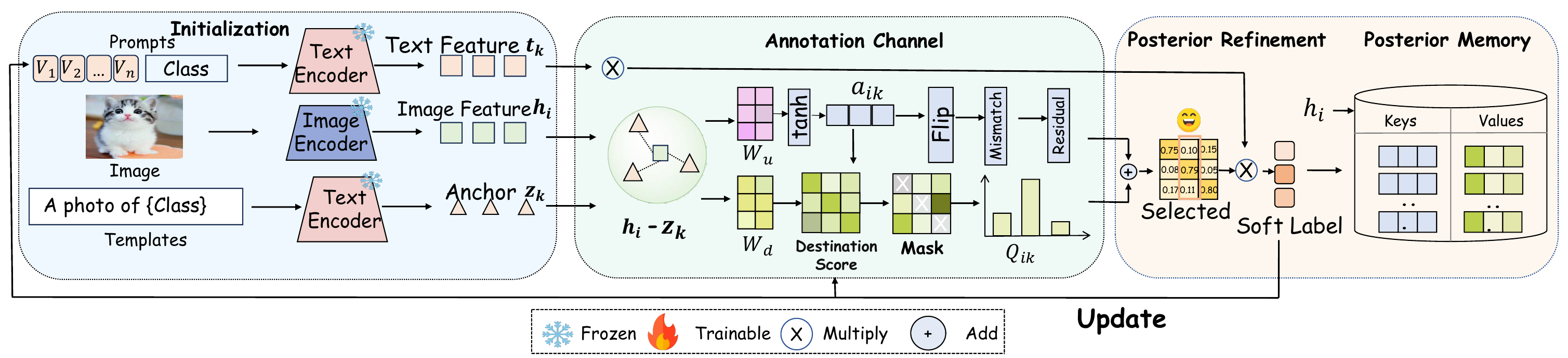}
    \caption{
    \textbf{Training framework of \texttt{SACP}.}
    \texttt{SACP} elegantly combines a prompt classifier with a structured
annotation channel to infer clean-label posteriors,
which guide prompt optimization through generalized
expectation-maximization.}
    \label{fig:overview}
\end{figure*}

However, real-world annotations often unavoidably contain label noise,
which can degrade prompt learning performance \cite{coprompt},
particularly in few-shot settings.
Existing approaches address this issue through robust
objectives \cite{coprompt}, sample selection \cite{robustclip}, label refinement \cite{Guo2024JoAPR}, or
cross-modal alignment \cite{na-mvp}.
Although effective, these methods primarily focus on mitigating unreliable supervision rather than modeling how observed annotations arise from latent clean classes.

We instead inspect this problem from \textbf{a different perspective}: Noisy annotations are not merely corrupted labels, but observations generated from latent clean classes. Modeling this annotation process allows us to estimate how likely each candidate clean class is to produce the observed label, thereby extracting useful supervision even from potentially corrupted annotations. As illustrated in Fig.~\ref{fig:teaser},
\texttt{SACP} models annotation generation to infer
clean-label posteriors, reducing semantic misalignment
caused by noisy annotations. Unlike a binary clean or noisy decision, this process captures both the probability of label corruption and the distribution of incorrect labels. Importantly, these patterns may depend on semantic relationships between classes and the visual characteristics of individual images. However, estimating such instance-dependent annotation patterns remains challenging under limited supervision. This motivates us to leverage the semantic knowledge of pretrained VLMs to model annotation generation and infer probabilistic clean-label distributions without relying on deterministic label correction.

Based on the above motivations and insights, we propose \texttt{SACP}, a Semantic-aware Annotation Channel-based Prompt Learning framework for robust VLM adaptation under noisy labels. \texttt{SACP} couples a prompt classifier that predicts latent clean classes with a structured annotation channel that models how these classes generate observed annotations. To estimate this channel under limited supervision, we elegantly combine fixed semantic priors from pretrained VLMs with instance-specific visual corrections to capture both class-level confusion patterns and sample-dependent variations. By integrating semantic predictions with annotation likelihoods, \texttt{SACP} infers probabilistic clean-label posteriors, which guide prompt and channel optimization through a generalized expectation-maximization procedure. Furthermore, we introduce a posterior-aware memory that stores image features and their inferred label distributions to refine predictions during inference.

Extensive experiments on five benchmark datasets under
symmetric and asymmetric label noise demonstrate the
effectiveness of SACP.
Notably, \texttt{SACP} outperforms the strongest baseline by
10.17 percentage points on OxfordPets under 50\%
asymmetric noise.
Our main contributions are summarized as follows:
\begin{itemize}[leftmargin=0.5cm]
    \item We provide a new perspective on noisy-label
    prompt learning by viewing noisy annotations as
    observations generated from latent clean classes,
    allowing useful supervision to be extracted from
    potentially corrupted labels.

    \item We propose SACP, which combines a structured
    annotation channel with semantic priors and
    instance-specific corrections.
    A generalized expectation-maximization procedure
    guides prompt optimization using clean-label
    posteriors, while a posterior-aware memory extends
    these distributions to inference.

    \item Extensive experiments across multiple benchmark
datasets demonstrate that \texttt{SACP} achieves
state-of-the-art performance under diverse
noise settings, highlighting its robustness
to noisy supervision.
\end{itemize}

\section{Preliminaries}
\label{sec:preliminaries}

Let $\mathcal{D}=\{(x_i,\tilde{y}_i)\}_{i=1}^{N}$
be a noisy few-shot training set, where $x_i$
is an image, $\tilde{y}_i\in\{1,\ldots,C\}$
is its observed label, and $C$ is the number
of classes. The clean label $y_i$ is unknown.
Our goal is to learn a classifier without clean
training labels or prior knowledge of label noise.
Following CoOp~\cite{CoOp}, we freeze the pretrained
visual and text encoders and optimize learnable
prompt tokens $\boldsymbol{\theta}$.
The prompt classifier predicts
$p_{ik}=p_{\boldsymbol{\theta}}(y_i=k\mid x_i)$
from image-text similarity.
We denote the normalized image feature by
$\mathbf{h}_i$.
An annotation channel
$T_{ikj}=T_{\boldsymbol{\phi}}(\tilde{y}_i=j
\mid y_i=k,x_i)$, parameterized by
$\boldsymbol{\phi}$, models the probability
of observing class $j$ given clean class $k$
and image $x_i$.

\section{Method}
\label{sec:method}

As illustrated in Fig.~\ref{fig:overview},
\texttt{SACP} elegantly combines semantic predictions with
an annotation channel to infer clean-label
posteriors for training and inference.
\S~\ref{sec:annotation_channel} models annotation
generation, \S~\ref{sec:posterior} introduces
posterior-guided optimization, and
\S~\ref{sec:memory} develops memory-assisted inference.

\subsection{Annotation Channel}
\label{sec:annotation_channel}
\begin{figure}[htbp]
    \centering
    \includegraphics[width=\columnwidth]{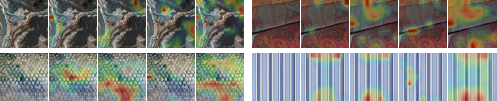}
    \caption{
    \textbf{Qualitative analysis on DTD.}
    Grad-CAM~\cite{grad-cam} compares noisy-label
    supervision, semantic evidence,
    annotation-channel attribution, and
    clean-versus-observed label contrast.
    }
    \label{fig:sacp-gradcam}
\end{figure}

We first introduce an annotation channel
to model semantic confusion and instance-dependent
label noise.
To support few-shot learning, the channel combines
fixed semantic anchors $\mathbf{z}_k$, derived from
handcrafted class-name templates, with learnable
visual corrections.
The mismatch between image features $\mathbf{h}_i$
and class anchors determines the label-flipping
probability, while anchor similarities guide
the distribution over incorrect classes.
Shared low-dimensional projections capture
instance-specific variations, and a uniform
component accounts for random corruption.
Fig.~\ref{fig:sacp-gradcam} visualizes the
semantic and annotation-channel evidence.

\subsection{Posterior Refinement}
\label{sec:posterior}

We then develop a posterior refinement strategy
to recover reliable supervision from noisy labels.
By integrating prompt predictions with annotation
likelihoods, we infer clean-label posteriors
using Bayes' rule:
\begin{equation}
\label{eq:latent_posterior}
q_{ik}
=
P(y_i=k\mid x_i,\tilde{y}_i)
=
\frac{p_{ik}T_{ik\tilde{y}_i}}
{\sum_{c=1}^{C}p_{ic}T_{ic\tilde{y}_i}}.
\end{equation}

\begin{figure}[htbp]
    \centering
    \includegraphics[width=\columnwidth]{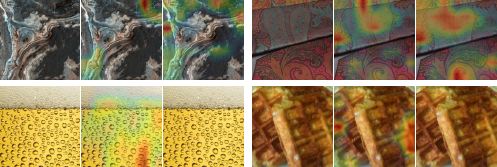}
    \caption{
    \textbf{Posterior disagreement.}
    Visualization of posterior disagreement
    for clean and corrupted annotations.
    }
    \label{fig:sacp-reliability-gradcam}
\end{figure}

Following generalized expectation-maximization,
we use the detached posteriors as soft supervision
to jointly optimize the prompt classifier and
annotation channel through weighted log-likelihoods.
An early regularization term prevents excessive
label flipping and gradually decays to zero,
avoiding hard label correction while retaining
uncertain samples.
Fig.~\ref{fig:sacp-reliability-gradcam}
illustrates posterior disagreement with
observed annotations.

\subsection{Posterior-Aware Memory}
\label{sec:memory}

Finally, we introduce a posterior-aware memory
to preserve the inferred clean-label information.
As shown in Fig.~\ref{fig:overview},
training image features $\mathbf{h}_i$ serve as keys,
while their detached posterior distributions
are stored as soft-label values.
This design associates visual representations
with refined label information rather than
the original noisy annotations.

\section{Experiments}

In this section, we present experimental details and empirical results results with analysis to evaluate the performance and effectiveness of \texttt{SACP} under different label noise patterns.

\subsection{Experimental Settings}

\textbf{Datasets and Metrics.}
We evaluate our method on six benchmark datasets, including five datasets with synthetically corrupted labels: Caltech101~\cite{caltech101}, DTD~\cite{dtd}, Flowers102~\cite{flowers102}, OxfordPets~\cite{oxfordpets}, and UCF101~\cite{ucf101-}, as well as Food101N~\cite{food101n}, which contains real-world noisy annotations.
Following NA-MVP~\cite{na-mvp}, we consider two noise patterns \cite{Co-teaching} for the five synthetic-noise benchmarks: symmetric noise (Sym.), where labels are uniformly replaced with other class labels, and asymmetric noise (Asym.), where labels are corrupted toward semantically similar classes.
For evaluation, we report top-1 classification accuracy on clean test sets, with higher accuracy indicating better classification performance under label noise.

\textbf{Baselines.}
Our comparison includes representative methods:
\begin{itemize}[leftmargin=0.5cm, itemsep=0pt]
    \item \textbf{CoOp}~\cite{CoOp} learns continuous
    text prompts while keeping the pretrained
    VLM backbone frozen.

    \item \textbf{CoOp+GCE}~\cite{gce} replaces
    the standard cross-entropy loss with generalized
    cross-entropy to reduce the influence of noisy labels.

    \item \textbf{JoAPR}~\cite{Guo2024JoAPR} uses
    adaptive thresholds to separate clean and noisy
    samples, followed by label refinement.

    \item \textbf{NLPrompt}~\cite{nlprompt} combines
    a robust MAE loss with optimal transport-based
    sample selection to mitigate label noise.

    \item \textbf{NA-MVP}~\cite{na-mvp} employs
    bi-directional multi-view prompt alignment
    and selective label refinement to improve
    robustness under noisy supervision.
\end{itemize}

\textbf{Implementation Details.}
Following the experimental protocol of CoOp~\cite{CoOp}, we use the same experimental configuration to ensure a fair comparison.
We optimize the model with SGD using an initial learning rate of 0.002, a momentum of 0.9, and a weight decay of $5\times10^{-4}$.
Unless stated otherwise, we employ ResNet-50 \cite{He2015DeepRL} as the image encoder and a Transformer \cite{transformer} containing 63M parameters as the text encoder.
For each dataset, experiments are conducted under the 16-shot setting, with performance evaluated on the complete clean test set.
The model is trained for 50 epochs by default.
We use 16 shared context tokens followed by the corresponding class token.
All reported results are averaged over three runs with different random seeds to reduce variability caused by randomness, and the highest accuracy is highlighted in bold.
All experiments are performed on a single NVIDIA RTX 3090 GPU.

\newcolumntype{C}{>{\centering\arraybackslash}m{0.82cm}}

\begin{table}[t]
    \centering
    \caption{
        Classification accuracy (\%) under different symmetric
        (Sym.) and asymmetric (Asym.). The best results are highlighted in bold.
    }
    \label{tab:noise_comparison}

    \setlength{\tabcolsep}{0.1pt}
    \renewcommand{\arraystretch}{0.95}

    \begin{tabular}{@{}l*{8}{C}@{}}
        \toprule

        \multirow{3}{*}{\textbf{Method}}
        & \multicolumn{8}{c}{\textbf{Noise Pattern}} \\

        \cmidrule(lr){2-9}

        & \multicolumn{6}{c}{\textbf{Sym.}}
        & \multicolumn{2}{c}{\textbf{Asym.}} \\

        & 0.125 & 0.25 & 0.375 & 0.5
        & 0.625 & 0.75 & 0.25 & 0.5 \\

        \midrule


        \rowcolor{gray!20}
        \multicolumn{9}{c}{\textbf{Caltech101}~\cite{caltech101}} \\
        \addlinespace[1pt]

        CoOp~\cite{CoOp}
        & 86.43 & 81.03 & 76.73 & 70.90
        & 61.33 & 46.90
        & 75.23 & 49.43 \\

        GCE~\cite{gce}
        & 92.00 & 90.90 & 90.80 & 89.30
        & 86.70 & 79.03
        & 91.20 & 85.80 \\

        JoAPR~\cite{Guo2024JoAPR}
        & 88.32 & 87.85 & 87.00 & 87.03
        & 84.55 & 80.15
        & 82.79 & 69.02 \\

        NLPrompt~\cite{nlprompt}
        & 91.73 & 91.13 & 90.77 & 89.93
        & 88.30 & 86.70
        & 91.17 & 89.27 \\

        NA-MVP~\cite{na-mvp}
        & 92.07 & 92.10 & 91.60 & 91.30
        & 90.07 & 89.37
        & 91.47 & 89.53 \\

        \rowcolor{blue!10}
        \textbf{SACP}
        & \textbf{92.44} & \textbf{92.77}
        & \textbf{92.17} & \textbf{91.59}
        & \textbf{91.78} & \textbf{90.21}
        & \textbf{92.21} & \textbf{90.42} \\

        \midrule


        \rowcolor{gray!20}
        \multicolumn{9}{c}{\textbf{DTD}~\cite{dtd}} \\
        \addlinespace[1pt]

        CoOp~\cite{CoOp}
        & 56.00 & 49.57 & 43.30 & 34.37
        & 27.83 & 17.27
        & 47.75 & 29.63 \\

        GCE~\cite{gce}
        & 61.00 & 59.83 & 56.80 & 50.73
        & 43.60 & 33.67
        & 57.57 & 43.97 \\

        JoAPR~\cite{Guo2024JoAPR}
        & 55.02 & 53.95 & 51.57 & 49.12
        & 44.24 & 35.90
        & 49.23 & 38.33 \\

        NLPrompt~\cite{nlprompt}
        & 62.97 & 61.23 & 59.17 & 55.17
        & 49.03 & 39.80
        & 60.60 & 50.80 \\

        NA-MVP~\cite{na-mvp}
        & 63.73 & 63.13 & 61.63 & 58.50
        & 52.93 & 48.63
        & 62.33 & 52.10 \\

        \rowcolor{blue!10}
        \textbf{SACP}
        & \textbf{66.13} & \textbf{65.20}
        & \textbf{63.22} & \textbf{58.81}
        & \textbf{54.82} & \textbf{50.30}
        & \textbf{63.24} & \textbf{56.57} \\

        \midrule


        \rowcolor{gray!20}
        \multicolumn{9}{c}{\textbf{Flowers102}~\cite{flowers102}} \\
        \addlinespace[1pt]

        CoOp~\cite{CoOp}
        & 88.93 & 83.50 & 77.93 & 70.10
        & 55.60 & 37.17
        & 74.70 & 42.60 \\

        GCE~\cite{gce}
        & 88.80 & 88.33 & 86.73 & 84.07
        & 78.37 & 70.37
        & 86.37 & 69.93 \\

        JoAPR~\cite{Guo2024JoAPR}
        & 84.90 & 84.70 & 79.75 & 77.13
        & 69.65 & 64.20
        & 79.57 & 55.47 \\

        NLPrompt~\cite{nlprompt}
        & 93.87 & 92.57 & 92.73 & 89.90
        & 84.77 & 76.80
        & \textbf{93.40} & 81.10 \\

        NA-MVP~\cite{na-mvp}
        & 94.20 & 93.30 & 92.00 & 90.47
        & 85.07 & 76.47
        & 91.37 & 78.43 \\

        \rowcolor{blue!10}
        \textbf{SACP}
        & \textbf{94.38} & \textbf{93.50}
        & \textbf{92.81} & \textbf{91.88}
        & \textbf{88.71} & \textbf{77.90}
        & 92.47 & \textbf{82.98} \\

        \midrule


        \rowcolor{gray!20}
        \multicolumn{9}{c}{\textbf{OxfordPets}~\cite{oxfordpets}} \\
        \addlinespace[1pt]

        CoOp~\cite{CoOp}
        & 76.50 & 66.73 & 60.33 & 47.03
        & 35.77 & 24.60
        & 66.20 & 38.73 \\

        GCE~\cite{gce}
        & 85.63 & 84.60 & 83.67 & 79.23
        & 71.40 & 53.17
        & 83.03 & 68.07 \\

        JoAPR~\cite{Guo2024JoAPR}
        & 83.17 & 82.05 & 80.62 & 79.05
        & 73.72 & 60.97
        & 76.82 & 62.85 \\

        NLPrompt~\cite{nlprompt}
        & 86.17 & 86.00 & 85.33 & 83.17
        & 80.03 & 70.77
        & 84.97 & 77.53 \\

        NA-MVP~\cite{na-mvp}
        & 88.50 & 88.40 & 88.23 & 88.13
        & 86.93 & 86.23
        & 87.53 & 79.33 \\

        \rowcolor{blue!10}
        \textbf{SACP}
        & \textbf{90.67} & \textbf{89.86}
        & \textbf{90.06} & \textbf{89.66}
        & \textbf{89.55} & \textbf{89.06}
        & \textbf{90.30} & \textbf{89.50} \\

        \midrule


        \rowcolor{gray!20}
        \multicolumn{9}{c}{\textbf{UCF101}~\cite{ucf101-}} \\
        \addlinespace[1pt]

        CoOp~\cite{CoOp}
        & 69.03 & 63.40 & 58.23 & 49.73
        & 40.83 & 26.30
        & 58.07 & 34.43 \\

        GCE~\cite{gce}
        & 74.00 & 73.63 & 72.57 & 69.37
        & 66.00 & 57.07
        & 71.87 & 67.97 \\

        JoAPR~\cite{Guo2024JoAPR}
        & 70.80 & 69.22 & 68.15 & 64.80
        & 61.82 & 57.52
        & 63.98 & 49.67 \\

        NLPrompt~\cite{nlprompt}
        & 74.83 & 73.40 & 72.83 & 70.33
        & 68.10 & 60.53
        & 73.58 & 65.97 \\

        NA-MVP~\cite{na-mvp}
        & 75.33 & 74.03 & 72.30 & 70.93
        & 68.43 & 63.93
        & 73.40 & 65.40 \\

        \rowcolor{blue!10}
        \textbf{SACP}
        & \textbf{77.33} & \textbf{75.78}
        & \textbf{75.49} & \textbf{72.39}
        & \textbf{71.15} & \textbf{68.24}
        & \textbf{75.33} & \textbf{70.87} \\

        \bottomrule

    \end{tabular}
\end{table}

\subsection{Performance Analysis}

\ding{182} \textbf{Performance Across Benchmark Datasets:} As shown in Tab.~\ref{tab:noise_comparison}, \texttt{SACP} achieves state-of-the-art performance across multiple benchmark datasets under diverse noise settings.
Compared with NA-MVP, \texttt{SACP} improves the average
performance by 2.10\% and 4.28\% under
symmetric and asymmetric noise, respectively.
The larger gains under asymmetric noise demonstrate
its stronger resistance to class-dependent label corruption.
Notably, \texttt{SACP} achieves a 10.17\% improvement
on OxfordPets at a 50\% asymmetric noise rate.
Its robustness also extends to severe symmetric noise,
with a 4.31\% gain on UCF101 at a 75\% noise rate.
These results highlight the effectiveness of SACP
in maintaining reliable predictions under noisy supervision.

\begin{table}[t]
    \centering
    \caption{
        Classification accuracy (\%) on Food101N
        under real-world label noise with different
        few-shot settings.
    }
    \label{tab:food101n}

    \setlength{\tabcolsep}{10pt}
    \renewcommand{\arraystretch}{0.8}

    \begin{tabular}{@{}lcccc@{}}
        \toprule

        \textbf{Method}
        & \textbf{4-shot}
        & \textbf{8-shot}
        & \textbf{16-shot}
        & \textbf{32-shot} \\

        \midrule

        NLPrompt~\cite{nlprompt}
        & 70.57 & 73.93 & 76.46 & 76.87 \\

        NA-MVP~\cite{na-mvp}
        & 76.10 & 76.27 & 76.90 & 77.03 \\

        \textbf{SACP}
        & \textbf{77.46}
        & \textbf{77.57}
        & \textbf{77.66}
        & \textbf{77.79} \\

        \bottomrule
    \end{tabular}
\end{table}

\noindent \ding{183} \textbf{Performance on Real-World Noisy Datasets:} As shown in Tab.~\ref{tab:food101n}, \texttt{SACP} consistently outperforms both baselines across all few-shot settings on Food101N \cite{food101n}. In particular, it achieves accuracies of 77.46\% and 77.57\% in the 4-shot and 8-shot settings, surpassing NA-MVP by 1.79\% and 1.70\%. Besides, \texttt{SACP} also maintains a performance advantage of 0.76 in both the 16-shot and 32-shot settings. These results demonstrate its effectiveness under real-world label noise, with particularly pronounced gains when labeled data are scarce.

\begin{figure}[htbp]
    \centering
    \includegraphics[width=0.9\columnwidth]{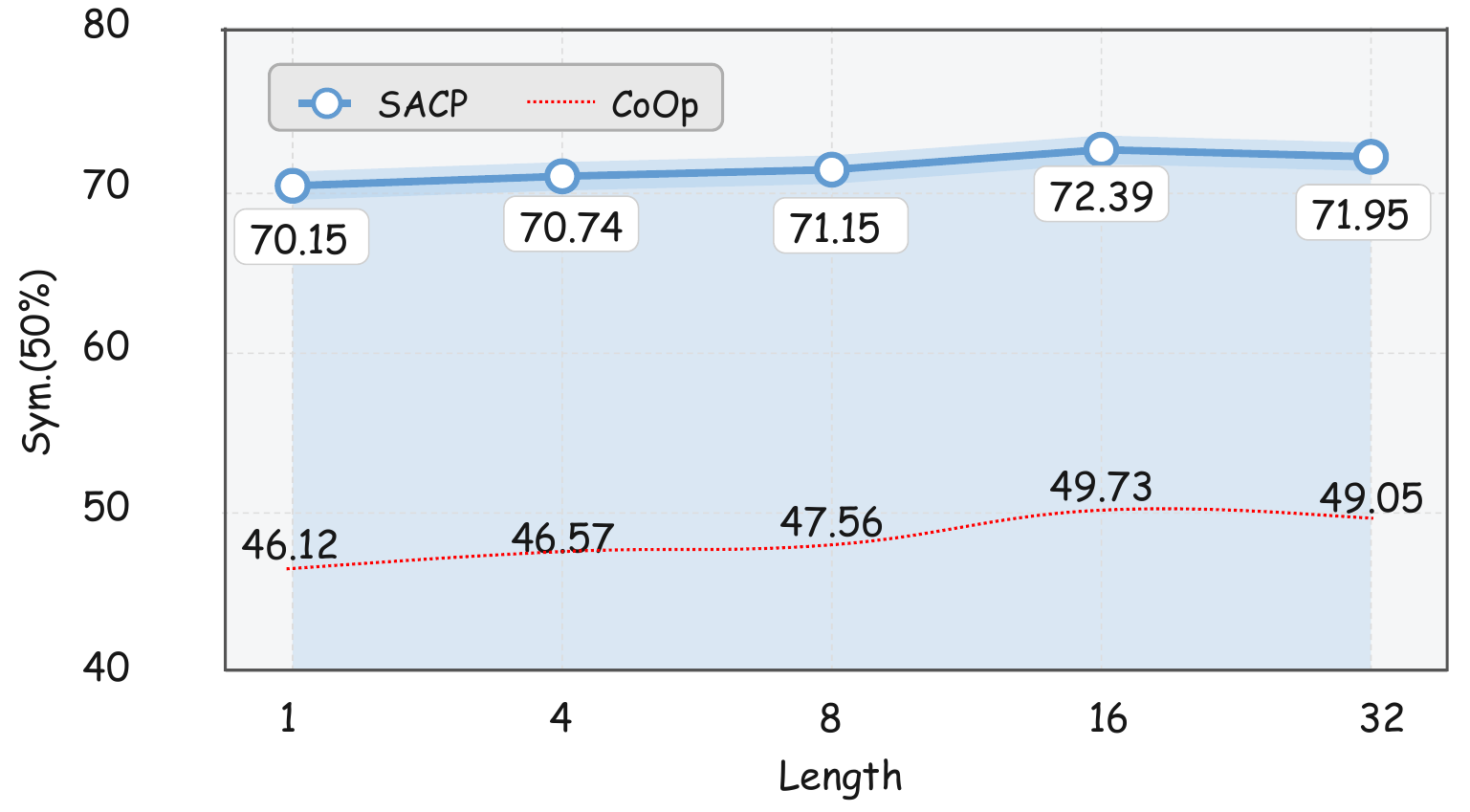}
    \caption{
    Performance with the different prompt lengths.}
    \label{fig:length}
\end{figure}

\subsection{Ablation Study}

\begin{table}[t]
\centering
\caption{Component ablation on DTD under 50\%
symmetric and asymmetric label noise.}
\label{tab:ablation}
\setlength{\tabcolsep}{5pt}
\renewcommand{\arraystretch}{1.12}
\resizebox{0.8\columnwidth}{!}{%
\begin{tabular}{lccc}
\toprule
\textbf{Method} & \textbf{Sym.} & \textbf{Asym.} & \textbf{Avg.} \\
\midrule
SACP
& \textbf{58.81} & \textbf{56.57}
& \textbf{57.69} \\
\midrule
w/o Semantic Channel
& 58.32 & 53.41 & 55.87 \\
w/o Semantic Constraint
& 57.48 & 53.82 & 55.65 \\
w/o Generalized EM
& 56.93 & 54.26 & 55.60 \\
w/o Posterior-aware Memory
& 57.36 & 54.79 & 56.08 \\
\bottomrule
\end{tabular}%
}
\end{table}

\ding{182} \textbf{Performance of different components:} Tab.~\ref{tab:ablation} presents an ablation study
on DTD under 50\% symmetric and asymmetric label noise.
Removing the semantic channel or semantic constraints
is expected to reduce performance, especially under
asymmetric noise, highlighting the importance of
semantic information in modeling annotation errors.
Replacing Generalized EM with marginal likelihood
optimization may reduce the effectiveness of
posterior learning.
Finally, removing posterior-aware memory may
degrade performance by discarding useful
posterior information during inference. These components work together to achieve the strong overall performance.

\noindent \ding{183} \textbf{Sensitivity of the prompt length:}
As shown in Fig.~\ref{fig:length}, \texttt{SACP} maintains stable performance across different prompt lengths. It consistently outperforms CoOp, which demonstrates its effectiveness.

\noindent \ding{184} \textbf{Sensitivity of the prompt shots:}
\begin{figure}[htbp]
    \centering
    \includegraphics[width=0.9\columnwidth]{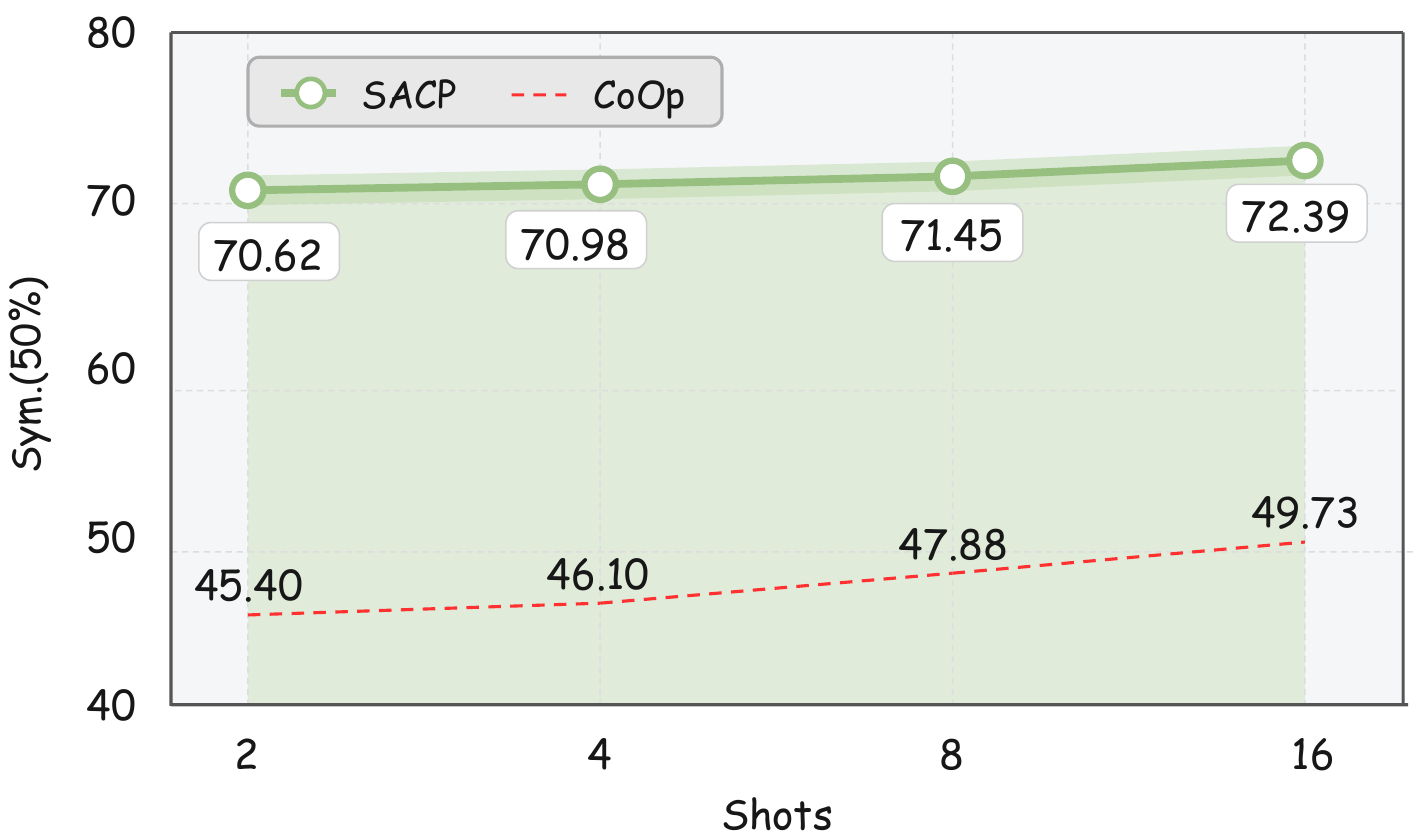}
    \caption{
    Performance with the different number of shots.}
    \label{fig:shots}
\end{figure}
As shown in Fig.~\ref{fig:shots}, \texttt{SACP} substantially outperforms CoOp across different shot settings, with gains ranging from 22.66\% to 25.22\%. Even with only 2 shots, \texttt{SACP} achieves 70.62\%, demonstrating its effectiveness with limited training examples.

\section{Conclusion}
\label{sec:conclusion}
In this work, we propose \texttt{SACP} for prompt learning with noisy few-shot supervision. \texttt{SACP} separates semantic recognition from annotation generation through a structured, instance-dependent annotation channel. Their joint evidence yields posterior soft targets for adaptation, while a posterior-aware memory supports inference. This formulation accounts for annotation uncertainty without requiring clean training labels or prior knowledge of the noise process.
Extensive experiments across diverse synthetic benchmarks and real-world noisy datasets demonstrate that \texttt{SACP} achieves state-of-the-art performance.
In future work, we plan to explore extending this annotation-channel paradigm to broader multi-modal robust web retrieval tasks.
\begin{acks}
To Robert, for the bagels and explaining CMYK and color spaces.
\end{acks}




\section*{Ethical Considerations}

\texttt{SACP} aims to improve prompt learning under noisy
supervision. However, semantic priors from
pretrained vision-language models may contain
biases, and posterior refinement may propagate
systematic annotation errors. Such biases could
lead to unequal classification performance across
different categories. To mitigate these risks,
we recommend evaluating class-level performance
and ensuring appropriate data usage practices.
For sensitive applications, human oversight
remains important to prevent unreliable
predictions from causing unintended harm.

\bibliographystyle{ACM-Reference-Format}
\bibliography{sample-base}

@String{Computer = "{IEEE} Computer" }

@String{Springer = "Springer-Verlag" }

@inproceedings{transformer,
  title={Attention Is All You Need},
  author={Ashish Vaswani and Noam Shazeer and Niki Parmar and Jakob Uszkoreit and Llion Jones and Aidan N. Gomez and Lukasz Kaiser and Illia Polosukhin},
  booktitle={Advances in Neural Information Processing Systems},
  volume={30},
  pages={5998--6008},
  year={2017},
  url={https://proceedings.neurips.cc/paper_files/paper/2017/hash/3f5ee243547dee91fbd053c1c4a845aa-Abstract.html}
}

@inproceedings{CLIP,
	title = {Learning transferable visual models from natural language supervision},
	isbn = {2640-3498},
	booktitle = {International conference on machine learning},
	publisher = {PMLR},
	author = {Radford, Alec and Kim, Jong Wook and Hallacy, Chris and Ramesh, Aditya and Goh, Gabriel and Agarwal, Sandhini and Sastry, Girish and Askell, Amanda and Mishkin, Pamela and Clark, Jack},
	year = {2021},
	pages = {8748--8763},
}

@inproceedings{nlprompt,
  title={NLPrompt: Noise-Label Prompt Learning for Vision-Language Models},
  author={Pan, B. and others},
  booktitle={Proceedings of the IEEE/CVF Conference on Computer Vision and Pattern Recognition (CVPR)},
  year={2025},
}

@inproceedings{Guo2024JoAPR,
  title={JoAPR: Cleaning the Lens of Prompt Learning for Vision-Language Models},
  author={Guo, Y. and others},
  booktitle={Proceedings of the IEEE/CVF Conference on Computer Vision and Pattern Recognition (CVPR)},
  year={2024},
}

@inproceedings{caltech101,
  title = {Learning Generative Visual Models from Few Training Examples: {{An}} Incremental Bayesian Approach Tested on 101 Object Categories},
  shorttitle = {Learning Generative Visual Models from Few Training Examples},
  booktitle = {2004 Conference on Computer Vision and Pattern Recognition Workshop},
  author = {{Fei-Fei}, Li and Fergus, Rob and Perona, Pietro},
  year = {2004},
  pages = {178--178},
  publisher = {IEEE},
  urldate = {2024-05-06}
}

@inproceedings{dtd,
  title = {Describing Textures in the Wild},
  booktitle = {Proceedings of the {{IEEE}} Conference on Computer Vision and Pattern Recognition},
  author = {Cimpoi, Mircea and Maji, Subhransu and Kokkinos, Iasonas and Mohamed, Sammy and Vedaldi, Andrea},
  year = {2014},
  pages = {3606--3613},
  urldate = {2024-05-04}
}

@inproceedings{flowers102,
  title = {Automated Flower Classification over a Large Number of Classes},
  booktitle = {2008 {{Sixth Indian}} Conference on Computer Vision, Graphics \& Image Processing},
  author = {Nilsback, Maria-Elena and Zisserman, Andrew},
  year = {2008},
  pages = {722--729},
  publisher = {IEEE},
  urldate = {2024-05-06}
}

@inproceedings{oxfordpets,
  title = {Cats and Dogs},
  booktitle = {2012 {{IEEE}} Conference on Computer Vision and Pattern Recognition},
  author = {Parkhi, Omkar M. and Vedaldi, Andrea and Zisserman, Andrew and Jawahar, C. V.},
  year = {2012},
  pages = {3498--3505},
  publisher = {IEEE},
  urldate = {2024-05-04}
}

@article{ucf101,
  title = {Two-Stream Collaborative Learning with Spatial-Temporal Attention for Video Classification},
  author = {Peng, Yuxin and Zhao, Yunzhen and Zhang, Junchao},
  year = {2018},
  journal = {IEEE Transactions on Circuits and Systems for Video Technology},
  volume = {29},
  number = {3},
  pages = {773--786},
  publisher = {IEEE},
  isbn = {1051-8215}
}

@inproceedings{food101n,
  title = {Cleannet: {{Transfer}} Learning for Scalable Image Classifier Training with Label Noise},
  booktitle = {Proceedings of the {{IEEE}} Conference on Computer Vision and Pattern Recognition},
  author = {Lee, Kuang-Huei and He, Xiaodong and Zhang, Lei and Yang, Linjun},
  year = {2018},
  pages = {5447--5456}
}

@article{CoOp,
	title = {Learning to prompt for vision-language models},
	volume = {130},
	number = {9},
	journal = {International Journal of Computer Vision},
	author = {Zhou, Kaiyang and Yang, Jingkang and Loy, Chen Change and Liu, Ziwei},
	year = {2022},
	note = {ISBN: 0920-5691
Publisher: Springer},
	pages = {2337--2348},
}

@article{He2015DeepRL,
  title={Deep Residual Learning for Image Recognition},
  author={Kaiming He and X. Zhang and Shaoqing Ren and Jian Sun},
  journal={2016 IEEE Conference on Computer Vision and Pattern Recognition (CVPR)},
  year={2015},
  pages={770-778},
  url={https://api.semanticscholar.org/CorpusID:206594692}
}

@misc{li2025surveystateartlarge,
      title={A Survey of State of the Art Large Vision Language Models: Alignment, Benchmark, Evaluations and Challenges}, 
      author={Zongxia Li and Xiyang Wu and Hongyang Du and Fuxiao Liu and Huy Nghiem and Guangyao Shi},
      year={2025},
      eprint={2501.02189},
      archivePrefix={arXiv},
      primaryClass={cs.CV},
      url={https://arxiv.org/abs/2501.02189}, 
}

@inproceedings{jia2021scaling,
  title={Scaling up visual and vision-language representation learning with noisy text supervision},
  author={Jia, Chao and Yang, Yinfei and Xia, Ye and Chen, Yi-Ting and Parekh, Zarana and Pham, Hieu and Le, Quoc and Sung, Yun-Hsuan and Li, Zhen and Duerig, Tom},
  booktitle={ICML},
  pages={4904--4916},
  year={2021},
  organization={PMLR}
}

@article{geng2026seeing,
  title={Seeing is Believing: Robust Vision-Guided Cross-Modal Prompt Learning under Label Noise},
  author={Geng, Zibin and Jiang, Xuefeng and Li, Jia and Li, Zheng and Wen, Tian and Wu, Lvhua and Sun, Sheng and Wang, Yuwei and Liu, Min},
  journal={arXiv preprint arXiv:2604.09532},
  year={2026}
}

@article{tbd,
  title={Representation Decorrelation Guided Robust Image Retrieval against Label Noise},
  author={Jiang, Xuefeng and Wen, Tian and Li, Peng and Wu, Lvhua and Sun, Sheng and Wang, Yuwei and Liu, Min},
  journal={IEEE Transactions on Big Data},
  year={2026},
  publisher={IEEE}
}

@inproceedings{gce,
author = {Zhang, Zhilu and Sabuncu, Mert R.},
title = {Generalized cross entropy loss for training deep neural networks with noisy labels},
year = {2018},
publisher = {Curran Associates Inc.},
address = {Red Hook, NY, USA},
booktitle = {Proceedings of the 32nd International Conference on Neural Information Processing Systems},
pages = {8792–8802},
numpages = {11},
location = {Montr\'{e}al, Canada},
series = {NIPS'18}
}

@inproceedings{blip,
author = {Li, Junnan and Li, Dongxu and Savarese, Silvio and Hoi, Steven},
title = {BLIP-2: bootstrapping language-image pre-training with frozen image encoders and large language models},
year = {2023},
publisher = {JMLR.org},
booktitle = {Proceedings of the 40th International Conference on Machine Learning},
articleno = {814},
numpages = {13},
location = {Honolulu, Hawaii, USA},
series = {ICML'23}
}

@misc{na-mvp,
      title={Noise-aware few-shot learning through bi-directional multi-view prompt alignment}, 
      author={Lu Niu and Cheng Xue},
      year={2026},
      eprint={2603.11617},
      archivePrefix={arXiv},
      primaryClass={cs.CV},
      url={https://arxiv.org/abs/2603.11617}, 
}

@misc{Co-teaching,
      title={Co-teaching: Robust Training of Deep Neural Networks with Extremely Noisy Labels}, 
      author={Bo Han and Quanming Yao and Xingrui Yu and Gang Niu and Miao Xu and Weihua Hu and Ivor Tsang and Masashi Sugiyama},
      year={2018},
      eprint={1804.06872},
      archivePrefix={arXiv},
      primaryClass={cs.LG},
      url={https://arxiv.org/abs/1804.06872}, 
}

@article{grad-cam,
   title={Grad-CAM: Visual Explanations from Deep Networks via Gradient-Based Localization},
   volume={128},
   ISSN={1573-1405},
   url={http://dx.doi.org/10.1007/s11263-019-01228-7},
   DOI={10.1007/s11263-019-01228-7},
   number={2},
   journal={International Journal of Computer Vision},
   publisher={Springer Science and Business Media LLC},
   author={Selvaraju, Ramprasaath R. and Cogswell, Michael and Das, Abhishek and Vedantam, Ramakrishna and Parikh, Devi and Batra, Dhruv},
   year={2019},
   month=Oct, pages={336–359} }

@Inbook{zero,
author="Zhou, Kaiyang
and Yang, Jingkang
and Loy, Chen Change
and Liu, Ziwei",
editor="Zhou, Kaiyang
and Liu, Ziwei
and Gao, Peng",
title="Differentiable Prompt Learning for Vision-Language Models",
bookTitle="Large Vision-Language Models: Pre-training, Prompting, and Applications",
year="2026",
publisher="Springer Nature Switzerland",
address="Cham",
pages="115--133",
isbn="978-3-031-94969-2",
doi="10.1007/978-3-031-94969-2_5",
url="https://doi.org/10.1007/978-3-031-94969-2_5"
}

@misc{plot,
      title={PLOT: Prompt Learning with Optimal Transport for Vision-Language Models}, 
      author={Guangyi Chen and Weiran Yao and Xiangchen Song and Xinyue Li and Yongming Rao and Kun Zhang},
      year={2023},
      eprint={2210.01253},
      archivePrefix={arXiv},
      primaryClass={cs.CV},
      url={https://arxiv.org/abs/2210.01253}, 
}

@misc{manu2,
      title={Multi-Modal Guided Multi-Source Domain Adaptation for Object Detection}, 
      author={Sangin Lee and Seokjun Kwon and Jeongmin Shin and Namil Kim and Yukyung Choi},
      year={2026},
      eprint={2605.13140},
      archivePrefix={arXiv},
      primaryClass={cs.CV},
      url={https://arxiv.org/abs/2605.13140}, 
}

@inproceedings{cocoop,
    title={Conditional Prompt Learning for Vision-Language Models},
    author={Zhou, Kaiyang and Yang, Jingkang and Loy, Chen Change and Liu, Ziwei},
    booktitle={IEEE/CVF Conference on Computer Vision and Pattern Recognition (CVPR)},
    year={2022}
}

@misc{fpl,
      title={Harmonizing Generalization and Personalization in Federated Prompt Learning}, 
      author={Tianyu Cui and Hongxia Li and Jingya Wang and Ye Shi},
      year={2024},
      eprint={2405.09771},
      archivePrefix={arXiv},
      primaryClass={cs.LG},
      url={https://arxiv.org/abs/2405.09771}, 
}

@misc{manu,
      title={Modeling Variants of Prompts for Vision-Language Models}, 
      author={Ao Li and Zongfang Liu and Xinhua Li and Jinghui Zhang and Pengwei Wang and Hu Wang},
      year={2025},
      eprint={2503.08229},
      archivePrefix={arXiv},
      primaryClass={cs.CV},
      url={https://arxiv.org/abs/2503.08229}, 
}

@misc{force,
      title={Florence: A New Foundation Model for Computer Vision}, 
      author={Lu Yuan and Dongdong Chen and Yi-Ling Chen and Noel Codella and Xiyang Dai and Jianfeng Gao and Houdong Hu and Xuedong Huang and Boxin Li and Chunyuan Li and Ce Liu and Mengchen Liu and Zicheng Liu and Yumao Lu and Yu Shi and Lijuan Wang and Jianfeng Wang and Bin Xiao and Zhen Xiao and Jianwei Yang and Michael Zeng and Luowei Zhou and Pengchuan Zhang},
      year={2021},
      eprint={2111.11432},
      archivePrefix={arXiv},
      primaryClass={cs.CV},
      url={https://arxiv.org/abs/2111.11432}, 
}

@misc{coprompt,
      title={TecoPrompt: Temporal-Conservative Prompt Learning for Vision-Language Models}, 
      author={Zeyi Shao and Haowen Hua and Jiaxin Zhang and John See and Zeyd Boukhers and Cong Yang},
      year={2026},
      eprint={2609.16858},
      archivePrefix={arXiv},
      primaryClass={cs.CV},
      url={https://arxiv.org/abs/2609.16858}, 
}

@misc{robustclip,
      title={Robust CLIP: Unsupervised Adversarial Fine-Tuning of Vision Embeddings for Robust Large Vision-Language Models}, 
      author={Christian Schlarmann and Naman Deep Singh and Francesco Croce and Matthias Hein},
      year={2024},
      eprint={2402.12336},
      archivePrefix={arXiv},
      primaryClass={cs.LG},
      url={https://arxiv.org/abs/2402.12336}, 
}

@misc{ucf101-,
      title={UCF101: A Dataset of 101 Human Actions Classes From Videos in The Wild}, 
      author={Khurram Soomro and Amir Roshan Zamir and Mubarak Shah},
      year={2012},
      eprint={1212.0402},
      archivePrefix={arXiv},
      primaryClass={cs.CV},
      url={https://arxiv.org/abs/1212.0402}, 
}










\end{document}